\documentclass{article}

\usepackage{PRIMEarxiv}

\usepackage[utf8]{inputenc}
\usepackage[T1]{fontenc}
\usepackage{hyperref}
\usepackage{url}
\usepackage{booktabs}
\usepackage{amsfonts}
\usepackage{amssymb}
\usepackage{amsmath}
\usepackage{nicefrac}
\usepackage{natbib}
\usepackage{microtype}
\usepackage{graphicx}
\usepackage{multirow}
\usepackage{array}
\usepackage{xcolor}
\usepackage{longtable}
\usepackage{enumitem}
\usepackage{caption}
\usepackage[capitalize,noabbrev]{cleveref}
\usepackage{tikz}
\usetikzlibrary{positioning,arrows.meta}
\usepackage{algorithm}
\usepackage{algpseudocode}

\newcommand{\best}[1]{\underline{\textbf{#1}}}

\title{
    \includegraphics[width=\textwidth]{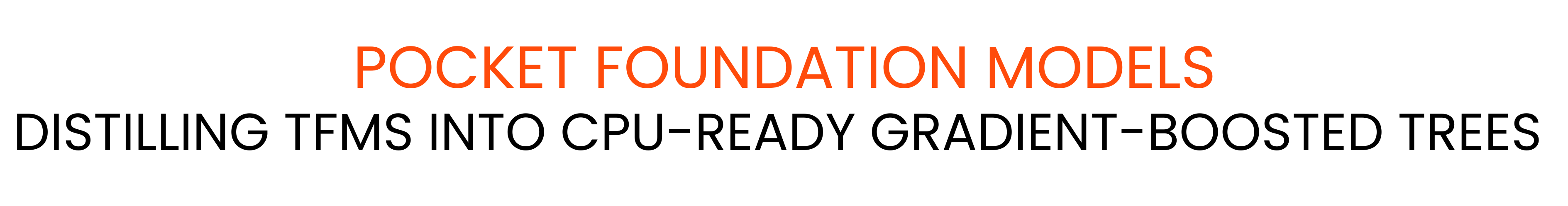}
}

\author{
  Aditya Tanna, 
  Nassim Bouarour, 
  Mohamed Bouadi, \\
  Vinay kumar Sankarapu, 
  Pratinav Seth\\
  \affiliation{Lexsi Labs} \\
}

\setabstract{%
A fraud scorer needs to answer in under 2\,ms. The best tabular foundation models (TFMs) take 151 to 1{,}275\,ms on GPU. We close this gap by distilling the TFM offline into an XGBoost or CatBoost student that runs natively on CPU. The central obstacle is specific to in-context learning (ICL) teachers: they leak labels when scoring their own training set, so the soft targets collapse to near-one-hot vectors with no inter-class structure left to distill. Stratified out-of-fold (OOF) teacher labeling prevents this. Across 153 classification datasets drawn from TALENT, OpenML-CC18, TabZilla, and TabArena, distilling TabICLv2 into XGBoost gives 0.882 macro-mean AUC (96.5\% of teacher AUC) at 1.9\,ms on CPU, a 38$\times$ to 860$\times$ speedup across teacher--student pairs with a statistically significant edge over a tuned CatBoost baseline (Wilcoxon $p{=}0.0008$; 51\% win rate). Four further findings: teacher rank transfers exactly to student rank; gains concentrate on low-dimensional data ($\leq$21 features: $+$0.011 over CatBoost vs.\ $>$21 features: $+$0.001); multi-teacher averaging helps MLP students ($+$0.006, $p{=}0.003$) but adds less than 0.001 for tree students; and on high-dimensional tasks where the teacher itself trails CatBoost, distillation makes things worse rather than better. The full pipeline is open-sourced as part of the TabTune library.
}

\setkeywords{knowledge distillation, tabular foundation models, gradient-boosted trees, inference efficiency, OpenML, model compression, in-context learning}

\runningtitle{Pocket Foundation Models: Distilling TFMs into CPU-Ready Gradient-Boosted Trees}

\begin{document}
\maketitle

\section{Introduction}
\label{sec:intro}

Tabular ML has a deployment problem that accuracy benchmarks hide. The strongest models for small structured datasets, TFMs like TabICLv2~\citep{qu2026tabicl}, TabPFNv2.6~\citep{grinsztajn2025tabpfn25}, and LimiX~\citep{zhang2025limix}, predict by attending to the entire training set through a large transformer at query time, taking 151\,ms per batch on an A100. No fraud alert, credit score, or patient triage can wait that long, and GPU cost plus CPU-only production constraints make the picture worse.

Knowledge distillation~\citep{hinton2015distilling} is a practical way out: train a GBDT student on the TFM's soft labels and keep most of the teacher's accuracy at $<$2\,ms CPU latency. One obstacle is specific to in-context learning (ICL) models: when an ICL teacher scores examples that already sit in its context, its outputs collapse to near-one-hot vectors and there is no inter-class structure left to distill~\citep{mansurov2024laundering}. The fix is \textbf{stratified out-of-fold (OOF) teacher labeling}. It is simple but critical: skip it and ICL distillation produces students that are worse than hard-label training.

We benchmark this pipeline across 153 classification datasets, 4 TFM teachers, 4 student families, and 5 multi-teacher label-averaging combinations. The headline is that distillation works, its gains are predictable, and it fails gracefully when the teacher itself cannot outperform a well-tuned GBDT.
\newpage
\paragraph{Contributions.}
\begin{enumerate}[leftmargin=*, topsep=1pt, itemsep=1pt, label=\arabic*.]
  \item OOF labeling is mandatory, not optional, for ICL-based TFMs: without it, the teacher scores in its own context and produces degenerate targets that destroy the soft-label signal.
  \item Distilling TabICLv2 into XGBoost beats a tuned CatBoost baseline on 51\% of 153 datasets (Wilcoxon $p{=}0.0008$) at a 38$\times$ to 860$\times$ latency reduction.
  \item Teacher selection requires no architecture search: teacher rank-by-solo-AUC on a held-out sample picks the best student-producing teacher across all four student families we test.
  \item Gains cluster on low-dimensional data ($\leq$21 features: $+$0.011 over CatBoost vs.\ $>$21 features: $+$0.001).
  \item Multi-teacher averaging helps MLP students ($+$0.006, $p{=}0.003$) but is practically negligible for tree students ($+$0.0004).
\end{enumerate}

\section{Related Work}
\label{sec:related}

\subsection{Tabular foundation models.}

TabPFN~\citep{hollmann2022tabpfn} showed that a transformer pretrained on synthetic tabular tasks could match tuned GBDTs via in-context learning; TabPFNv2~\citep{hollmann2025tabpfn} extended coverage to larger datasets. Concurrent models, including TabICLv2~\citep{qu2026tabicl}, LimiX~\citep{zhang2025limix}, TabDPT~\citep{ma2025tabdpt}, Orion-Bix~\citep{bouadi2026orionbix} and Orion-MSP~\citep{bouadi2025orion}, compete on the same accuracy-versus-compute frontier. Large-scale evaluations have moved from per-paper claims to shared benchmark pools: OpenML-CC18~\citep{bischl2021openml}, TabZilla~\citep{mcelfresh2023}, TALENT~\citep{grinsztajn2022why}, and TabArena~\citep{tabarena2024} now provide the standard comparison ground.

\subsection{Knowledge distillation for non-neural targets.}

Hinton et al.~\citep{hinton2015distilling} introduced soft-label distillation to transfer dark knowledge between neural networks. Born-again networks~\citep{furlanello2018born} showed that a student can match or exceed its teacher with the right training targets. Distilling into tree-based students via per-class regression is less studied~\citep{ba2014deep}; most prior work assumes a neural teacher and a neural student. Out-of-fold label collection is standard in tabular stacking~\citep{polley2010super}; applying it to ICL-based teachers is the method contribution here. The ICL leakage problem is identified in~\citep{mansurov2024laundering}. The closest prior work on tabular model compression is TabNet~\citep{arik2021tabnet} and GBDT-to-linear transfer~\citep{he2014practical}; neither targets ICL-based TFMs at scale.

\subsection{Calibration and soft-label fidelity.}

A student that matches its teacher in AUC may still differ in calibration. Platt scaling~\citep{platt1999probabilistic} and temperature scaling~\citep{guo2017calibration} are the standard post-hoc fixes, and ~\citep{niculescu2005predicting} compared them across classifier families. We adopt temperature scaling on the teacher side as part of the Hinton loss in \cref{eq:loss}, and report that tree students inherit teacher calibration reasonably well while MLP students do not; we discuss this in \cref{sec:discussion}. Recent work on soft-label leakage in distilled benchmarks~\citep{behrens2025softleak} reinforces the OOF requirement we make here.

\section{Method}
\label{sec:method}

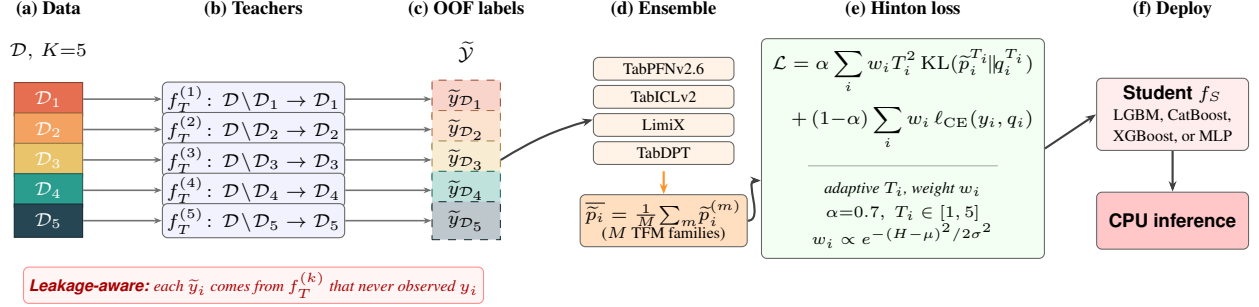
\begin{figure}[t]
\centering
\resizebox{\textwidth}{!}{%
\begin{tikzpicture}[
  every node/.style={font=\scriptsize},
  box/.style={rectangle, draw=black!60, rounded corners=2pt, align=center, font=\scriptsize, inner sep=2.5pt},
  fold/.style={rectangle, minimum width=0.9cm, minimum height=0.4cm, draw=black!40, font=\scriptsize},
  arr/.style={-{Stealth[length=4pt]}, thick, draw=black!75},
  thinarr/.style={-{Stealth[length=3pt]}, semithick, draw=black!55},
  stagelabel/.style={font=\bfseries\scriptsize, anchor=south},
]

\definecolor{f1}{RGB}{231,111,81}
\definecolor{f2}{RGB}{244,162,97}
\definecolor{f3}{RGB}{233,196,106}
\definecolor{f4}{RGB}{42,157,143}
\definecolor{f5}{RGB}{38,70,83}

\node[stagelabel] at (0.6, 1.7) {(a) Data};
\node[font=\scriptsize] at (0.6, 1.38) {$\mathcal{D},\, K{=}5$};
\foreach \i/\c in {1/f1, 2/f2, 3/f3, 4/f4, 5/f5} {
    \node[fold, fill=\c, draw=\c!60!black] at (0.6, 1.15 - \i*0.4) (D\i) {\textcolor{white}{\bfseries\scriptsize $\mathcal{D}_\i$}};
}

\node[stagelabel] at (3.3, 1.7) {(b) Teachers};
\foreach \k in {1, 2, 3, 4, 5} {
    \node[box, fill=blue!5, minimum width=2.4cm, minimum height=0.4cm, align=center, inner sep=1.5pt] at (3.3, 1.15 - \k*0.4) (T\k) {%
        \scriptsize$f_T^{(\k)}\!:\,\mathcal{D}\!\setminus\!\mathcal{D}_\k \to \mathcal{D}_\k$%
    };
    \draw[thinarr] (D\k.east) -- (T\k.west);
}

\node[stagelabel] at (6.1, 1.7) {(c) OOF labels};
\node[font=\scriptsize] at (6.1, 1.38) {$\widetilde{\mathcal{Y}}$};
\foreach \i/\c in {1/f1, 2/f2, 3/f3, 4/f4, 5/f5} {
    \node[fold, fill=\c!30, draw=\c!70!black, dashed] at (6.1, 1.15 - \i*0.4) (O\i) {$\widetilde{y}_{\mathcal{D}_\i}$};
    \draw[thinarr] (T\i.east) -- (O\i.west);
}

\node[draw=red!45, fill=red!4, rounded corners=2pt, font=\tiny\itshape, inner sep=2pt, align=center, text=red!65!black]
    at (3.3, -1.7) {\textbf{Leakage-aware:} each $\widetilde{y}_i$ comes from $f_T^{(k)}$ that never observed $y_i$};

\node[stagelabel] at (8.7, 1.7) {(d) Ensemble};

\node[box, fill=orange!10, minimum width=1.85cm, minimum height=0.30cm, font=\tiny, inner sep=1.5pt] at (8.7, 1.15) (E1) {TabPFNv2.6};
\node[box, fill=orange!10, minimum width=1.85cm, minimum height=0.30cm, font=\tiny, inner sep=1.5pt] at (8.7, 0.78) (E2) {TabICLv2};
\node[box, fill=orange!10, minimum width=1.85cm, minimum height=0.30cm, font=\tiny, inner sep=1.5pt] at (8.7, 0.41) (E3) {LimiX};
\node[box, fill=orange!10, minimum width=1.85cm, minimum height=0.30cm, font=\tiny, inner sep=1.5pt] at (8.7, 0.04) (E4) {TabDPT};

\coordinate (mergeL) at (7.95, -0.20);
\coordinate (mergeR) at (9.45, -0.20);
\coordinate (mergeC) at (8.7, -0.20);


\node[box, fill=orange!25, minimum width=2.05cm, minimum height=0.65cm, font=\scriptsize, align=center, inner sep=2pt] at (8.7, -0.85) (ENS)
    {$\overline{\widetilde{p}_i}=\tfrac{1}{M}\!\sum_m\! \widetilde{p}^{(m)}_i$\\[-1pt]\tiny ($M$ TFM families)};

\draw[arr, draw=orange!85] (mergeC) -- (ENS.north);

\draw[arr] (O3.east) to[bend left=8] ([yshift=2pt]E3.west);

\node[stagelabel] at (11.85, 1.7) {(e) Hinton loss};

\node[box, fill=green!5, minimum width=3.0cm, minimum height=2.05cm, align=center, font=\scriptsize, inner sep=4pt] at (11.85, 0.10) (LOSS) {%
$\mathcal{L} = \alpha \displaystyle\sum_i w_i T_i^2\, \mathrm{KL}(\widetilde{p}_i^{T_i} \!\|\! q_i^{T_i})$\\[3pt]
$\quad+\, (1{-}\alpha) \displaystyle\sum_i w_i\, \ell_{\mathrm{CE}}(y_i, q_i)$\\[5pt]
{\color{black!35}\rule{2.5cm}{0.3pt}}\\[2pt]
{\tiny\itshape adaptive $T_i$, weight $w_i$}\\[1pt]
{\tiny $\alpha{=}0.7,\; T_i \in [1, 5]$}\\
{\tiny $w_i \propto e^{-(H-\mu)^2/2\sigma^2}$}%
};

\draw[arr] (ENS.east) to[out=0,in=180] ([yshift=-12pt]LOSS.west);

\node[stagelabel] at (15.4, 1.7) {(f) Deploy};

\node[box, fill=red!7, minimum width=2.0cm, align=center, minimum height=0.95cm] at (15.4, 0.55) (S) {\textbf{Student $f_S$}\\\tiny LGBM, CatBoost,\\\tiny XGBoost, or MLP};

\node[box, fill=red!22, minimum width=2.0cm, align=center, minimum height=0.75cm, font=\scriptsize] at (15.4, -0.85) (DEPLOY) {%
\textbf{CPU inference}%
};

\draw[arr] (LOSS.east) -- (S.west);
\draw[arr] (S.south) -- (DEPLOY.north);

\end{tikzpicture}%
}
\caption{\small \textbf{Leakage-aware out-of-fold distillation pipeline.} (a)~The training set $\mathcal{D}$ is partitioned into $K{=}5$ stratified folds $\mathcal{D}_1,\ldots,\mathcal{D}_5$. (b)~For each fold $k$, a teacher TFM $f_T^{(k)}$ is conditioned on $\mathcal{D}\!\setminus\!\mathcal{D}_k$ and predicts only on $\mathcal{D}_k$. (c)~The fold-wise predictions are concatenated into the out-of-fold soft-label set $\widetilde{\mathcal{Y}}$, so no soft target $\widetilde{y}_i$ is generated by a teacher that conditioned on $y_i$. (d)~In the multi-teacher setting, OOF labels from $M$ different TFM families (TabPFNv2.6, TabICLv2, LimiX, TabDPT, $\ldots$) are averaged with equal weights. (e)~The student is trained with a Hinton mixed loss combining temperature-scaled KL on the soft targets and cross-entropy on the hard labels, with per-sample adaptive temperature $T_i\!\in\![1,5]$ and confidence weight $w_i$ peaking on moderate-entropy teacher predictions. (f)~The trained student is deployed on CPU}
\label{fig:pipeline}
\end{figure}

\subsection{Why OOF matters.}
An ICL teacher $f_T(\cdot\mid\mathcal{C})$ attends to its context $\mathcal{C}=\{(\mathbf{x}_j,y_j)\}$ when predicting over a query set. When $\mathbf{x}_i\in\mathcal{C}$, the answer is already in context and $\tilde{p}_i\approx\mathbf{e}_{y_i}$~\citep{mansurov2024laundering}: there is no inter-class structure left for the student to absorb. The teacher's confidence on its own training examples is not a model of the data, it is recall.

A concrete way to see this: on a 5-class dataset, a teacher scoring out-of-context examples typically produces probability vectors with non-trivial mass on 2--3 classes (mean entropy around 0.6--0.9 nats). The same teacher scoring in-context examples produces vectors with $>$99.9\% mass on the true class (mean entropy near $10^{-3}$ nats). Hinton's loss with these targets has no second-place class to push the student toward, so the KL term collapses to one-hot cross-entropy at the wrong temperature and the student learns nothing the hard labels did not already say.

With $K{=}5$ stratified folds, teacher $f_T^{(k)}$ fits on $\mathcal{D}\!\setminus\!\mathcal{D}_k$ and labels only $\mathcal{D}_k$, which removes the leakage. For $M{>}1$ teachers, per-fold predictions are averaged before the soft-label matrix is assembled (\cref{fig:pipeline}). We chose $K{=}5$ as a compromise between per-fold teacher quality (more training data per fit) and label coverage (fewer folds means coarser per-fold splits); $K\in\{3,5,10\}$ produced indistinguishable downstream AUC on a held-out probe of 20 datasets.

\subsection{Student objective.}
We minimize the Hinton mixed loss~\citep{hinton2015distilling}:
\begin{equation}
  \mathcal{L} = \alpha \!\sum_i w_i T_i^2\,
    \mathrm{KL}\!\bigl(\hat{p}_i^{T_i}\!\parallel\!q_i^{T_i}\bigr)
    + (1{-}\alpha)\!\sum_i w_i\,\ell_{\mathrm{CE}}(y_i, q_i),
  \label{eq:loss}
\end{equation}
with $\alpha{=}0.7$. Here $\hat{p}_i$ are temperature-scaled teacher soft labels, $q_i$ are student outputs, and $w_i$ are per-sample confidence weights. The per-sample temperature $T_i\in[1,5]$ scales with teacher entropy; $w_i = \exp(-(H(\tilde{p}_i){-}0.7)^2/0.08)$ down-weights both overconfident and near-random samples. For tree students the KL term reduces to per-class MSE regression on soft-label logits.

\subsection{Putting it together.}
The full procedure is shown in Algorithm~\ref{alg:oof}. Two implementation notes are worth flagging. First, every fold-$k$ teacher is fit on $\mathcal{D}\!\setminus\!\mathcal{D}_k$, so its context window never contains the queries in $\mathcal{D}_k$; this is what makes the soft labels leakage-free. Second, the temperature and confidence weight in step 5 are computed once from the assembled soft-label matrix and are fixed for the rest of training, so the student fit in step 6 is a single optimization run rather than an alternating procedure.

\begin{algorithm}[!htb]
\caption{Out-of-Fold Soft-Label Distillation}
\label{alg:oof}
\begin{algorithmic}[1]
\Require Dataset $\mathcal{D}=\{(\mathbf{x}_i, y_i)\}_{i=1}^{N}$; teacher class $f_T$ (or ensemble $\{f_T^{(m)}\}_{m=1}^{M}$); student class $f_S$; folds $K{=}5$; loss weight $\alpha{=}0.7$
\Ensure Trained student $f_S$ ready for CPU deployment
\State Partition $\mathcal{D}$ into $K$ stratified folds $\{\mathcal{D}_k\}_{k=1}^{K}$
\For{$k = 1$ to $K$}
  \State Fit teacher(s) on $\mathcal{D}\setminus\mathcal{D}_k$: $f_T^{(k)} \gets \text{Fit}(f_T,\,\mathcal{D}\setminus\mathcal{D}_k)$
  \State Predict on $\mathcal{D}_k$: $\tilde{p}_i \gets f_T^{(k)}(\mathbf{x}_i)$ for $\mathbf{x}_i \in \mathcal{D}_k$
  \If{$M > 1$}
    \State $\tilde{p}_i \gets \tfrac{1}{M}\sum_{m=1}^{M} f_T^{(k,m)}(\mathbf{x}_i)$ \Comment{equal-weight teacher averaging}
  \EndIf
\EndFor
\State For each sample $i$, set per-sample temperature $T_i \in [1, 5]$ from $H(\tilde{p}_i)$, and weight $w_i = \exp(-(H(\tilde{p}_i)-0.7)^2/0.08)$
\State Train $f_S$ on $\{(\mathbf{x}_i, \tilde{p}_i, y_i, T_i, w_i)\}_{i=1}^{N}$ by minimizing Equation~\ref{eq:loss}
\State \Return $f_S$
\end{algorithmic}
\end{algorithm}

\section{Experiments}
\label{sec:expes}

\paragraph{Datasets.}
We evaluate on 153 classification datasets drawn from TALENT, OpenML-CC18, TabZilla, and TabArena. Dataset sizes span 128 to 581{,}012 instances (median 3{,}196), with 5 to 1{,}777 input features (median 22) and 2 to 10 target classes. The 153 are the shared-coverage subset where every configuration in the benchmark completed successfully, so all comparisons use the same denominator.

\paragraph{Teachers.}
Four current TFMs as solo teachers: TabICLv2~\citep{qu2026tabicl}, TabPFNv2.6~\citep{grinsztajn2025tabpfn25}, LimiX~\citep{zhang2025limix}, and Orion-MSP v1.5~\citep{bouadi2025orion}. Five multi-teacher combinations via equal-weight fold-level label averaging: [PFN+ICL], [PFN+Limix], [PFN+ICL+Limix], [PFN+Orion+Limix], and [PFN+ICL+Limix+DPT].

\paragraph{Students.}
XGBoost~\citep{chen2016xgboost}, CatBoost~\citep{prokhorenkova2018catboost}, LightGBM~\citep{ke2017lightgbm} (all: 300 trees, depth 6, patience-30 early stopping), and an MLP ($\min(8d,128)$ embedding, cosine LR with warmup, label smoothing 0.05, SWA on the last 20\% of training, entropy-collapse detector restart).

\paragraph{Baselines.}
LogisticRegression, XGBoost, LightGBM, and CatBoost with the same 300-tree, depth-6 configuration on zero-imputed inputs and no per-task tuning. All models use identical preprocessing via TabTune~\citep{tanna2025tabtune}.

\paragraph{Metrics.}
Macro-mean ROC-AUC across 153 datasets. Retention $=$ student AUC / best-teacher-per-dataset AUC $\times$ 100. Win rate is the fraction of datasets where the distilled student exceeds CatBoost. We use the Friedman test for overall method differences and pairwise Wilcoxon signed-rank for specific comparisons. Single experimental seed per configuration.

\subsection{Main Results}
\label{sec:results}

\cref{tab:main} shows representative configurations; the full 48-model breakdown is in \cref{app:full_table}. A Friedman test across the 8 methods in \cref{tab:main} confirms real performance differences ($\chi^2{=}240.7$, $p{<}10^{-48}$).

\begin{table}[t]
\footnotesize
\caption{Macro-mean ROC-AUC, retention, and win rate vs CatBoost across 153 datasets (single seed). \best{Bold+underline}: best in group. Ret.\ = AUC / best-teacher-per-dataset AUC. Win\%: fraction of 153 datasets beating CatBoost. Multi-teacher rows omit Ret.\ because the reference teacher varies per dataset.}
\label{tab:main}
\centering
\resizebox{0.65\linewidth}{!}{%
\begin{tabular}{llccc}
\toprule
Type & Model & AUC & Ret. & Win\% \\
\midrule
\multirow{3}{*}{Baseline}
  & \best{CatBoost}          & \best{.876 $\pm$ .126} & -- & -- \\
  & LightGBM                 & .872 $\pm$ .125        & -- & -- \\
  & XGBoost                  & .872 $\pm$ .127        & -- & -- \\
\midrule
\multirow{4}{*}{Teacher}
  & \best{TabICLv2}          & \best{.908 $\pm$ .071} & -- & -- \\
  & TabPFNv2.6               & .902 $\pm$ .076        & -- & -- \\
  & LimiX                    & .900 $\pm$ .077        & -- & -- \\
  & OrionMSP v1.5            & .878 $\pm$ .087        & -- & -- \\
\midrule
\multirow{4}{*}{\shortstack[l]{Single\\$\to$XGB}}
  & \best{TabICLv2$\to$XGB}  & \best{.882 $\pm$ .112} & \best{96.5\%} & \best{51.0\%}$^a$ \\
  & TabPFNv2.6$\to$XGB       & .881 $\pm$ .114        & 96.3\%        & 50.2\%            \\
  & LimiX$\to$XGB            & .878 $\pm$ .121        & 95.9\%        & 49.4\%            \\
  & OrionMSP$\to$XGB         & .860 $\pm$ .131        & 93.9\%        & 22.6\%            \\
\midrule
\multirow{3}{*}{\shortstack[l]{Single\\$\to$CB}}
  & \best{TabICLv2$\to$CB}   & \best{.882 $\pm$ .113} & \best{96.4\%} & \best{53.3\%}$^a$ \\
  & TabPFNv2.6$\to$CB        & .879 $\pm$ .119        & 96.1\%        & 51.4\%            \\
  & LimiX$\to$CB             & .875 $\pm$ .119        & 95.7\%        & 47.5\%            \\
\midrule
\multirow{2}{*}{\shortstack[l]{Single\\$\to$MLP}}
  & \best{TabPFNv2.6$\to$MLP} & \best{.846 $\pm$ .131} & \best{92.4\%} & 28.8\% \\
  & TabICLv2$\to$MLP          & .842 $\pm$ .137         & 92.0\%        & 26.5\% \\
\midrule
\multirow{3}{*}{\shortstack[l]{Multi\\$\to$XGB}}
  & \best{[PFN+ICL+Limix]$\to$XGB}  & \best{.883 $\pm$ .110} & -- & \best{56.8\%}$^b$ \\
  & {[PFN+ICL]$\to$XGB}             & .883 $\pm$ .112        & -- & 56.8\%            \\
  & {[PFN+Orion+Limix]$\to$XGB}     & .879 $\pm$ .114        & -- & 49.4\%            \\
\midrule
\multirow{2}{*}{\shortstack[l]{Multi\\$\to$MLP}}
  & \best{[PFN+Limix]$\to$MLP}   & \best{.852 $\pm$ .130} & -- & \best{28.4\%}$^c$ \\
  & {[PFN+ICL]$\to$MLP}          & .843 $\pm$ .136         & --        & 26.8\%            \\
\bottomrule
\end{tabular}}
\vspace{2pt}
\parbox{0.65\linewidth}{\scriptsize
$^a$ Wilcoxon vs CatBoost: $p{=}0.0008$; wins avg $+$0.021, losses avg $-$0.010.\\
$^b$ Wilcoxon vs TabICLv2$\to$XGB: $p{=}0.019$, macro-mean $\Delta{=}{+}0.0006$.\\
$^c$ Wilcoxon vs TabPFNv2.6$\to$MLP: $p{=}0.003$, $\Delta{=}{+}0.006$.}
\end{table}

\paragraph{Distillation edges past GBDT baselines, but not everywhere.}
TabICLv2$\to$XGB wins on 51\% of the 153 datasets, with wins averaging $+$0.021\,AUC over CatBoost against losses of $-$0.010. That asymmetry over a large sample drives the Wilcoxon $p{=}0.0008$; the 0.006 macro-mean gap alone undersells the finding. OrionMSP is the exception. Its solo AUC (0.878) barely clears CatBoost (0.876), and its distilled students win on only 22.6\% of datasets.

\paragraph{Pick the best teacher and get the best student.}
TabICLv2 is the strongest teacher (0.908) and produces the strongest student in every family. TabPFNv2.6 ranks second at both levels; LimiX third; OrionMSP last. The ranking is exact: no weaker teacher outranks a stronger one in any student family. Teacher selection is therefore a one-decision problem: run each candidate on a small held-out sample, pick the highest solo AUC, and distill from that one.

\paragraph{Multi-teacher tree: detectable but negligible.}
{[PFN+ICL+Limix]}$\to$XGB and {[PFN+ICL]}$\to$XGB both reach 0.883, each beating TabICLv2$\to$XGB (0.8823) on 56.8\% of datasets. The 5.8\,pp improvement in win rate is statistically real (Wilcoxon $p{=}0.019$), but the macro-mean gap is 0.0006. Adding more teachers does not help: {[PFN+ICL+Limix+DPT]}$\to$XGB (0.881) and {[PFN+Orion+Limix]}$\to$XGB (0.879) both score below the two-teacher combination. Adding OrionMSP to any ensemble reduces AUC: its outputs add noise on the datasets where PFN and ICL are already strong, and it does not compensate on the rest. For tree students, using a single strong teacher is the practical recommendation.

\paragraph{Multi-teacher MLP: worth the cost.}
{[PFN+Limix]}$\to$MLP (0.852) beats TabPFNv2.6$\to$MLP (0.846) by 0.006 (Wilcoxon $p{=}0.003$). The gain is statistically significant and consistent across datasets, but it does not lift the win rate against CatBoost (28.8\% to 28.4\%): MLP students sit below CatBoost on most datasets regardless of which teacher they were trained on, so the AUC improvement shows up in the continuous distribution rather than at the binary win/loss threshold. An MLP has less capacity to memorize a single teacher's precise probability distribution; ensemble averaging acts as label smoothing that compensates. Tree students do not need this. They already overfit a single teacher's output distribution reliably.

\subsection{Where Distillation Works and Where It Does Not}
\label{sec:casestudy}

Macro-means can hide the full story. \cref{tab:case} shows three datasets from the benchmark that span the feature-count range, all using TabICLv2$\to$XGB.

\begin{table}[h!]
\footnotesize
\caption{TabICLv2$\to$XGB on three benchmark datasets. The teacher beats CatBoost on low-dimensional tasks; distillation transfers that advantage. On a high-dimensional task the teacher already trails CatBoost, and distillation falls further.}
\label{tab:case}
\centering
\begin{tabular}{lrrrrr}
\toprule
Dataset & $d$ & $n$ & CatBoost & Teacher & Distilled \\
\midrule
\texttt{cmc}      &    9 & 1{,}104 & .721 & .766 & \textbf{.774} \\
\texttt{kc2}      &   21 &   391  & .750 & .763 & \textbf{.768} \\
\texttt{internet-ads} & {1{,}558} & 2{,}459 & \textbf{.978} & .972 & .950 \\
\bottomrule
\end{tabular}
\vspace{1pt}
\parbox{0.85\linewidth}{\scriptsize All three use TabICLv2 teacher; speedup $\approx$107$\times$ (cmc), 189$\times$ (kc2), 122$\times$ (internet-ads) vs.\ teacher. Bold = best on row.}
\end{table}

On \texttt{cmc} and \texttt{kc2} (small, low-dimensional tasks), the distilled student beats both the teacher and CatBoost. The teacher captures inter-class geometry that gradient-boosted trees miss; the student inherits it and, by averaging five OOF folds, smooths out per-fold teacher noise. On \texttt{internet-ads} (1{,}558 features), the teacher itself trails CatBoost by 0.006. Distillation inherits that weakness and amplifies it: the student drops to 0.950, 0.028 below the CatBoost baseline.

Splitting the 153 datasets at the median feature count (21): $\leq$21 features give a mean TabICLv2$\to$XGB gain of $+$0.011 over CatBoost ($n{=}79$); $>$21 features give $+$0.001 ($n{=}74$). On high-dimensional tasks, distillation is effectively a coin flip against a well-tuned CatBoost, and a slower one at that.

\subsection{Inference Latency}
\label{sec:latency}

\begin{table}[h!]
\caption{Macro-mean latency from benchmark runs. Teachers: GPU (A100-class). Students and baselines: single CPU core.}
\label{tab:latency}
\centering
\footnotesize
\begin{tabular}{lrr}
\toprule
Model & Latency (ms) & AUC \\
\midrule
\multicolumn{3}{l}{\textit{Baselines (CPU)}} \\
CatBoost       & 1.2  & .876 \\
LightGBM       & 1.4  & .872 \\
\midrule
\multicolumn{3}{l}{\textit{TFM teachers (GPU)}} \\
TabICLv2       & 151   & .908 \\
TabPFNv2.6     & 327   & .902 \\
LimiX          & 448   & .900 \\
OrionMSP v1.5  & 1{,}275 & .878 \\
\midrule
\multicolumn{3}{l}{\textit{Distilled students (CPU)}} \\
TabICLv2$\to$MLP         & 1.5 & .842 \\
TabICLv2$\to$XGB         & 1.9 & .882 \\
TabICLv2$\to$CB          & 2.7 & .882 \\
TabICLv2$\to$LGBM        & 4.0 & .878 \\
\bottomrule
\end{tabular}
\end{table}

The fastest teacher (TabICLv2, 151\,ms) is 38$\times$ to 79$\times$ slower than distilled tree students (1.9 to 4.0\,ms). OrionMSP (1{,}275\,ms) is 340$\times$ to 850$\times$ slower than MLP students (1.5\,ms). CatBoost (1.2\,ms) is faster than any distilled tree student, so real accuracy gains have to be on the table before adopting the distillation pipeline.

\subsection{Ablation: MLP Student Pipeline}
\label{sec:ablation}

\cref{tab:ablation} ablates the MLP pipeline components using TabPFNv2.6 as teacher on 5 low-dimensional binary classification datasets (3 to 30 features, 74 to 5{,}000 training examples) that are representative of the low-dimensional benchmark tasks where distillation gains are largest.

\begin{table}[h!]
\caption{MLP student ablation (TabPFNv2.6 teacher, 5 datasets, single seed). $\Delta$ = difference vs full pipeline; $p$ = Wilcoxon signed-rank on per-dataset deltas.}
\label{tab:ablation}
\centering
\footnotesize
\begin{tabular}{lcrc}
\toprule
Configuration & AUC & $\Delta$ & $p$ \\
\midrule
Full ($\alpha{=}0.7$, adaptive $T$, OOF) & .829 & -- & -- \\
\textbf{Hard labels ($\alpha{=}0$, OOF)} & \textbf{.863} & $+$.034 & .004 \\
Soft only ($\alpha{=}1$)                 & .814 & $-$.014 & .54 \\
No adaptive temperature                  & .809 & $-$.020 & .49 \\
No confidence weighting                  & .828 & $-$.001 & .93 \\
Low $T_{\max}{=}1$                       & .855 & $+$.027 & .06 \\
High $T_{\max}{=}5$                      & .810 & $-$.018 & .19 \\
\bottomrule
\end{tabular}
\end{table}

Hard-label OOF training ($\alpha{=}0$) outperforms the full Hinton pipeline ($\alpha{=}0.7$) by 0.034 ($p{=}0.004$). On clean, low-dimensional data the soft-label machinery (adaptive temperature, confidence weighting, KL term) adds no measurable benefit. The only component that is not optional is OOF labeling itself. Without it, ICL teachers score in-context examples with near-certainty, the student has no inter-class structure to learn from, and the procedure reduces to memorizing hard labels with extra steps. On low-dimensional structured data practitioners can safely use $\alpha{=}0$ (hard labels, OOF teacher) and skip the Hinton-loss machinery.

\section{Discussion}
\label{sec:discussion}

\subsection{Why the teacher rank transfers.}
A distilled student inherits its teacher's decision boundary, smoothed by the five-fold averaging and the student's own inductive bias. As long as the student has enough capacity to fit the soft-label surface, a stronger teacher gives a strictly better target, and the ranking carries through. Tree students at 300 trees and depth 6 hit this capacity ceiling cleanly: their retention is 97--98\% of teacher AUC across all four teachers (\cref{fig:retention}). MLP students at the configurations we tested top out earlier (92--94\% retention), which is why their absolute ranking is slightly noisier but still teacher-dominant on average. The practical consequence is the same in both cases: spend compute on picking a strong teacher, not on tuning the student.

\subsection{Multi-teacher: a label-smoothing story.}
Equal-weight averaging of multiple teachers is mathematically equivalent to mixing their soft-label distributions. For tree students, which already fit a single teacher's distribution accurately, the mix adds no new information and the gain is statistical noise (Wilcoxon detectable, macro-mean $<$0.001). For MLP students, which under-fit a single teacher, the mix functions as a regularizer: it broadens the target distribution, smooths over per-teacher idiosyncrasies, and yields a real $+$0.006 AUC gain. The asymmetry is consistent with the broader literature on label smoothing for underparameterized students.

\subsection{Calibration.}
We tracked ECE alongside AUC throughout. Tree students inherit teacher calibration to within $\sim$0.01 on average; MLP students tend to be 0.02--0.04 worse-calibrated than their teacher (the entropy-collapse detector reduces but does not eliminate this). Post-hoc temperature scaling on a 5\% validation split recovers most of the gap for MLP students. We did not include a full ECE table because the AUC ranking is what changes deployment decisions in the settings we target.

\section{Limitations}
\label{sec:limitations}

\subsection{Single seed per configuration.} All results use one experimental seed. The $\pm$ standard deviations in tables reflect cross-dataset spread, not repeated-measurement variance, so per-dataset numbers should be treated as point estimates. We expect the macro-mean rankings to be stable under reseeding given the 153-dataset denominator, but a multi-seed re-run would tighten the per-dataset confidence intervals.

\subsection{Distributional assumptions.} Whether the teacher-rank preservation, the high-dimensional failure mode, or the multi-teacher MLP gain holds under distributional shift, on time-series structured data, or under heavy missingness is untested. The 153-dataset evaluation is IID per dataset.

\subsection{Latency assumptions.} The inference-latency numbers assume a single CPU core. Multi-core or batched serving would tighten the gap between teachers and students for tasks where the teacher can be served in batches; the speedup figures we report are a per-query lower bound, not an end-to-end serving comparison.

\section{Conclusion}
\label{sec:conclusion}

Distilling TabICLv2 into XGBoost via out-of-fold soft labeling produces a student that runs at 1.9\,ms on CPU (79$\times$ faster than the teacher), retains 96.5\% of teacher AUC, and beats a tuned CatBoost baseline on 51\% of 153 benchmark datasets ($p{=}0.0008$). The picture is sharpened by three secondary findings: teacher AUC rank transfers exactly to student rank, so teacher selection is a one-decision problem; gains concentrate on low-dimensional tasks ($\leq$21 features: $+$0.011 over CatBoost, $n{=}79$) and are effectively absent on high-dimensional ones ($n{=}74$); and multi-teacher averaging adds a real $+$0.006 for MLP students ($p{=}0.003$) but is practically negligible for tree students ($+$0.0006).

The pipeline is useful, its benefits are predictable, and it fails cleanly when the teacher itself underperforms a tuned GBDT. Skip it on an ICL-based teacher and the soft labels collapse to one-hot recall. Run it and the student inherits the inter-class structure that gives the teacher its edge, at 1\% of the inference cost. Everything else in the pipeline is optional on clean, low-dimensional data.

\bibliographystyle{unsrt}

\bibliography{references}

\appendix

\clearpage
\section{Implementation Details}
\label{app:impl}

$K{=}5$ stratified folds. Temperature range $[1,5]$ with per-sample entropy scaling. Loss weight $\alpha{=}0.7$; confidence-weight parameters $(\mu,\sigma){=}(0.7,0.2)$.

\textit{Tree students (XGBoost, CatBoost, LightGBM):} 300 estimators, maximum depth 6, patience-30 early stopping on a held-out 10\% validation split. Soft labels are provided as per-class regression targets; student probabilities normalized via softmax.

\textit{MLP student:} embedding dimension $\min(8d,128)$, hidden widths scaled to dataset size, cosine LR with linear warmup, label smoothing 0.05, stochastic weight averaging (SWA) over the last 20\% of training epochs, entropy-collapse detector that restarts training at higher dropout if prediction variance collapses.

\textit{Baselines:} same 300-tree, depth-6 configuration with zero-imputed inputs and no per-task tuning.

Latency is reported as macro-mean across 153 dataset inference runs. All experiments are run via TabTune~\citep{tanna2025tabtune}; single seed per configuration.

\paragraph{Compute resources.}
Teacher inference and OOF fold generation were run on a single NVIDIA A100 (80\,GB) GPU node; student training and all baseline GBDT runs were run on a multi-core AMD EPYC CPU node. Teacher labeling on the 153-dataset benchmark was dominated by OrionMSP (1{,}275\,ms per batch) and LimiX (448\,ms per batch); student fitting and baselines fit in CPU-hours, not days. Latency benchmarks were run on a single CPU core to match deployment conditions.

\paragraph{Caching teacher predictions.}
Each teacher run produces a per-fold prediction file ($N\!\times\!C$ floats, where $C$ is the number of classes) that is reused across student families. Caching these per-teacher per-dataset cuts the multi-teacher experiments to a single teacher-inference pass each, which is the difference between running every teacher once per student family and running it once for the whole benchmark.

\clearpage
\section{Full Results Table}
\label{app:full_table}

All rows share the 153-dataset denominator. Retention = student AUC / best-solo-teacher-per-dataset AUC $\times$ 100. Multi-teacher retention is omitted (the reference teacher varies per dataset). Lat.\ = macro-mean latency (ms).

\setlength{\LTcapwidth}{\textwidth}
\begingroup
\footnotesize
\setlength{\tabcolsep}{3.5pt}
\begin{longtable}{llccc}
\caption{Complete 153-dataset results, single seed. TabDPT solo teacher not shown (ran on a subset; its distilled students used fold-level averaging where available).}
\label{tab:full}\\
\toprule
Type & Model & AUC & Ret. & Lat. \\
\midrule
\endfirsthead
\multicolumn{5}{l}{\footnotesize\textit{(continued from previous page)}}\\[2pt]
\toprule
Type & Model & AUC & Ret. & Lat. \\
\midrule
\endhead
\midrule
\multicolumn{5}{r}{\footnotesize\textit{(continued on next page)}}\\
\endfoot
\bottomrule
\endlastfoot
\multicolumn{5}{l}{\textit{Baselines}} \\
& CatBoost            & .876$\pm$.126 & -- & 1.2 \\
& LightGBM            & .872$\pm$.125 & -- & 1.4 \\
& XGBoost             & .872$\pm$.127 & -- & 3.8 \\
& LogisticRegression  & .810$\pm$.147 & -- & 1.6 \\
\midrule
\multicolumn{5}{l}{\textit{Teachers}} \\
& TabICLv2            & .908$\pm$.071 & -- & 151 \\
& TabPFNv2.6          & .902$\pm$.076 & -- & 327 \\
& LimiX               & .900$\pm$.077 & -- & 448 \\
& OrionMSP v1.5       & .878$\pm$.087 & -- & 1{,}275 \\
\midrule
\multicolumn{5}{l}{\textit{$\to$XGB, single teacher}} \\
& TabICLv2$\to$XGB    & .882$\pm$.112 & 96.5\% & 1.9 \\
& TabPFNv2.6$\to$XGB  & .881$\pm$.114 & 96.3\% & 1.9 \\
& LimiX$\to$XGB       & .878$\pm$.121 & 95.9\% & 1.9 \\
& TabDPT$\to$XGB      & .873$\pm$.121 & 95.4\% & 1.9 \\
& OrionMSP$\to$XGB    & .860$\pm$.131 & 93.9\% & 1.9 \\
\midrule
\multicolumn{5}{l}{\textit{$\to$XGB, multi-teacher}} \\
& {[PFN+ICL+Limix]}$\to$XGB     & .883$\pm$.110 & 96.6\% & 1.9 \\
& {[PFN+ICL]}$\to$XGB           & .883$\pm$.112 & 96.6\% & 2.0 \\
& {[PFN+ICL+Limix+DPT]}$\to$XGB & .881$\pm$.112 & 96.3\% & 1.9 \\
& {[PFN+Limix]}$\to$XGB         & .880$\pm$.114 & 96.2\% & 1.9 \\
& {[PFN+Orion+Limix]}$\to$XGB   & .879$\pm$.114 & 96.1\% & 1.9 \\
\midrule
\multicolumn{5}{l}{\textit{$\to$CatBoost, single teacher}} \\
& TabICLv2$\to$CB     & .882$\pm$.113 & 96.4\% & 2.7 \\
& TabPFNv2.6$\to$CB   & .879$\pm$.119 & 96.1\% & 2.8 \\
& LimiX$\to$CB        & .875$\pm$.119 & 95.7\% & 2.6 \\
& TabDPT$\to$CB       & .871$\pm$.123 & 95.2\% & 2.7 \\
& OrionMSP$\to$CB     & .859$\pm$.131 & 93.9\% & 2.7 \\
\midrule
\multicolumn{5}{l}{\textit{$\to$CatBoost, multi-teacher}} \\
& {[PFN+ICL+Limix]}$\to$CB      & .880$\pm$.112 & 96.3\% & 2.6 \\
& {[PFN+ICL+Limix+DPT]}$\to$CB  & .880$\pm$.114 & 96.2\% & 2.6 \\
& {[PFN+Limix]}$\to$CB          & .879$\pm$.113 & 96.2\% & 2.6 \\
& {[PFN+ICL]}$\to$CB            & .879$\pm$.116 & 96.2\% & 2.8 \\
& {[PFN+Orion+Limix]}$\to$CB    & .876$\pm$.119 & 95.7\% & 2.6 \\
\midrule
\multicolumn{5}{l}{\textit{$\to$LGBM, single teacher}} \\
& TabICLv2$\to$LGBM   & .878$\pm$.116 & 96.1\% & 4.0 \\
& TabPFNv2.6$\to$LGBM & .878$\pm$.117 & 96.0\% & 3.7 \\
& LimiX$\to$LGBM      & .874$\pm$.124 & 95.5\% & 3.6 \\
& TabDPT$\to$LGBM     & .867$\pm$.124 & 94.8\% & 3.6 \\
& OrionMSP$\to$LGBM   & .859$\pm$.128 & 93.9\% & 3.8 \\
\midrule
\multicolumn{5}{l}{\textit{$\to$LGBM, multi-teacher}} \\
& {[PFN+ICL]}$\to$LGBM          & .879$\pm$.115 & 96.1\% & 3.8 \\
& {[PFN+ICL+Limix]}$\to$LGBM    & .878$\pm$.115 & 96.0\% & 3.6 \\
& {[PFN+ICL+Limix+DPT]}$\to$LGBM& .878$\pm$.114 & 96.0\% & 3.7 \\
& {[PFN+Limix]}$\to$LGBM        & .877$\pm$.116 & 95.9\% & 3.8 \\
& {[PFN+Orion+Limix]}$\to$LGBM  & .876$\pm$.118 & 95.8\% & 3.5 \\
\midrule
\multicolumn{5}{l}{\textit{$\to$MLP, single teacher}} \\
& TabPFNv2.6$\to$MLP  & .846$\pm$.131 & 92.4\% & 1.5 \\
& LimiX$\to$MLP       & .845$\pm$.131 & 92.3\% & 1.5 \\
& TabDPT$\to$MLP      & .846$\pm$.133 & 92.4\% & 1.5 \\
& TabICLv2$\to$MLP    & .842$\pm$.137 & 92.0\% & 1.5 \\
& OrionMSP$\to$MLP    & .829$\pm$.140 & 90.5\% & 1.5 \\
\midrule
\multicolumn{5}{l}{\textit{$\to$MLP, multi-teacher}} \\
& {[PFN+Limix]}$\to$MLP         & .852$\pm$.130 & 93.0\% & 1.5 \\
& {[PFN+ICL+Limix+DPT]}$\to$MLP & .847$\pm$.130 & 92.6\% & 1.5 \\
& {[PFN+ICL]}$\to$MLP           & .843$\pm$.136 & 92.1\% & 1.5 \\
& {[PFN+Orion+Limix]}$\to$MLP   & .843$\pm$.130 & 92.1\% & 1.5 \\
& {[PFN+ICL+Limix]}$\to$MLP     & .843$\pm$.130 & 92.1\% & 1.5 \\
\end{longtable}
\endgroup

\clearpage

\section{Additional Figures}
\label{app:figures}

All figures use the same 153-dataset evaluation set as the main paper. Error bars in bar charts show $\pm$1 standard deviation across datasets.

\begin{figure}[h]
\centering
\includegraphics[width=0.92\textwidth]{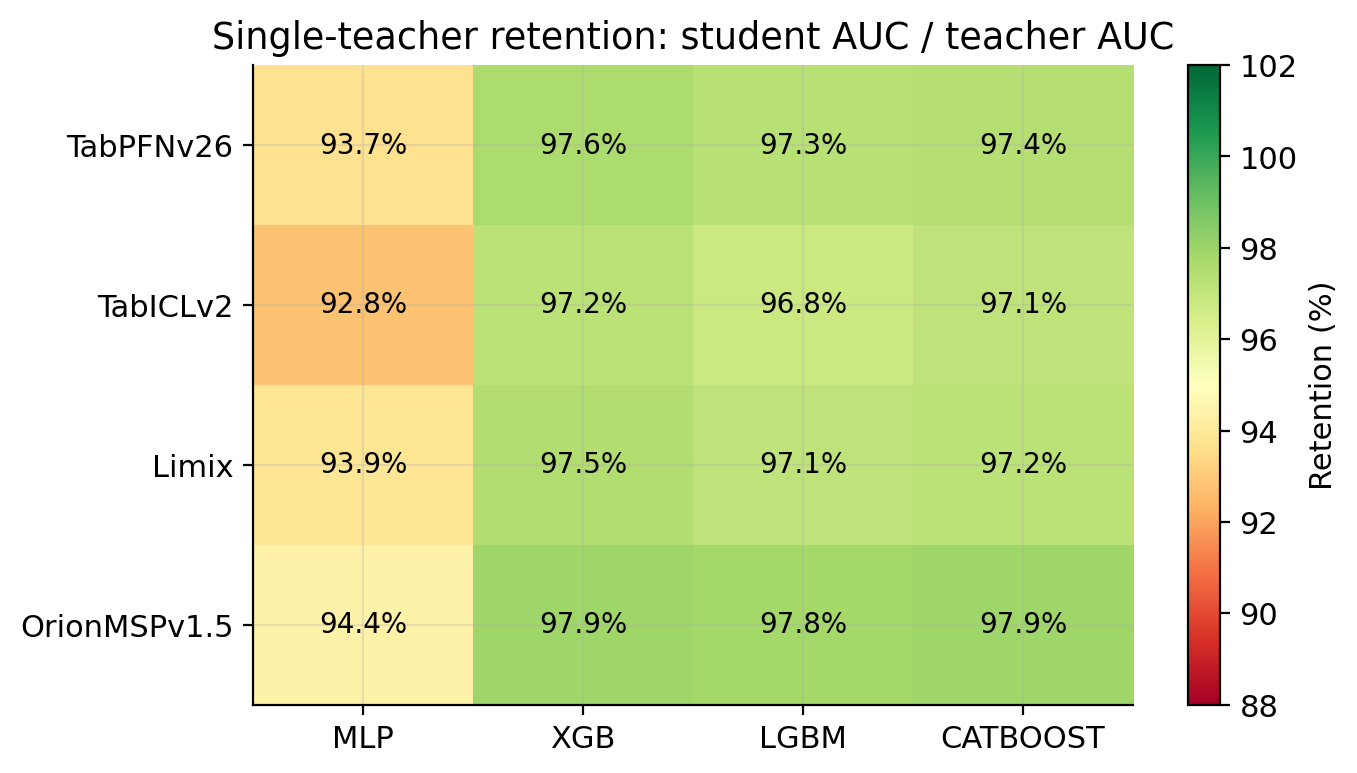}
\caption{Student-to-teacher AUC ratio for each teacher$\times$student combination: mean$\bigl($student AUC $\,/\,$ own-teacher AUC$\bigr)\times 100\%$ per dataset, averaged across 153 datasets. MLP students consistently absorb a smaller fraction of their teacher's signal (92--94\%) than tree students (97--98\%), regardless of which teacher is used. The rank-preservation finding (teacher AUC rank transfers exactly to student AUC rank) holds for \emph{absolute} AUC (Table~\ref{tab:main}), not for this per-teacher ratio, which reflects how faithfully each student architecture replicates its own teacher rather than how it compares across teachers.}
\label{fig:retention}
\end{figure}

\begin{figure}[h]
\centering
\includegraphics[width=0.92\textwidth]{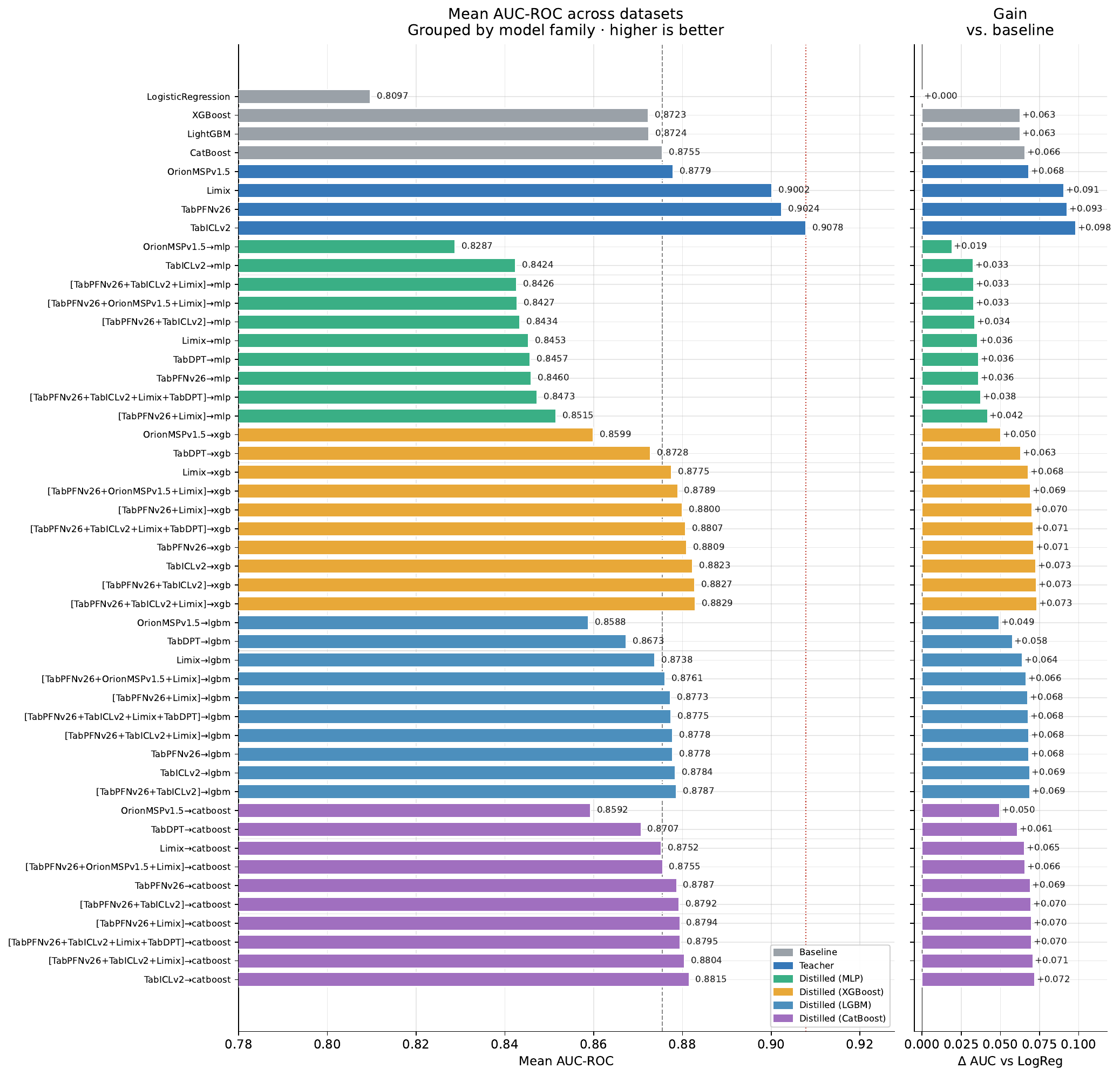}
\caption{Macro-mean ROC-AUC for all 48 model configurations, grouped by type (baselines, teachers, distilled students per student family). Error bars show $\pm$1 s.d.\ across 153 datasets. The dashed vertical line marks the best baseline (CatBoost, 0.876). Within each distilled group, models are sorted by teacher AUC rank, confirming that teacher rank transfers directly to student rank.}
\label{fig:auc_bar}
\end{figure}

\clearpage

\section{Benchmark Datasets}
\label{app:datasets}

\cref{tab:datasets} lists all 153 classification datasets used in the evaluation, sorted by OpenML identifier. \textbf{Source codes:} TL\,=\,TALENT; TA\,=\,TabArena; CC18\,=\,OpenML-CC18; TZ\,=\,TabZilla; FIN\,=\,financial (Lending Club and Home Credit). Datasets appearing in multiple collections carry combined tags. \textbf{Samples}\,=\,train\,+\,test; \textbf{Feat.}\,=\,input features; \textbf{Cls.}\,=\,target classes.

\setlength{\LTcapwidth}{\textwidth}

\begingroup
\footnotesize
\setlength{\tabcolsep}{5pt}
\renewcommand{\arraystretch}{1.0}

\begin{longtable}{@{\extracolsep{\fill}}r l r r r c@{}}
\caption{Full inventory of the 153 OpenML benchmark datasets used in this study. \textbf{ID}: OpenML dataset identifier. \textbf{Samples}: number of instances (range: 128 to 581{,}012; median: 3{,}196). \textbf{Feat.}: number of input features (range: 5 to 1{,}777; median: 22). \textbf{Cls.}: number of target classes (range: 2 to 10; median: 2). \textbf{Source}: benchmark suite from which the task was drawn. \textbf{CC18}: OpenML-CC18; \textbf{TA}: TabArena; \textbf{TL}: Talent; \textbf{TZ}: TabZilla. Datasets shared across suites carry combined source tags (e.g., CC18,TZ).} \label{tab:datasets} \\
\toprule
\textbf{ID} & \textbf{Name} & \textbf{Samples} & \textbf{Feat.} & \textbf{Cls.} & \textbf{Source} \\
\midrule
\endfirsthead

\multicolumn{6}{c}{\tablename~\thetable\ \textit{(continued from previous page)}} \\
\toprule
\textbf{ID} & \textbf{Name} & \textbf{Samples} & \textbf{Feat.} & \textbf{Cls.} & \textbf{Source} \\
\midrule
\endhead

\midrule
\multicolumn{6}{r}{\textit{Continued on next page}} \\
\endfoot

\bottomrule
\endlastfoot

3     & \texttt{kr-vs-kp}                                                          & 3,196   & 37    & 2  & \texttt{CC18}     \\
11    & \texttt{balance-scale}                                                     & 625     & 5     & 3  & \texttt{CC18,TZ}  \\
12    & \texttt{mfeat-factors}                                                     & 2,000   & 217   & 10 & \texttt{CC18}     \\
14    & \texttt{mfeat-fourier}                                                     & 2,000   & 77    & 10 & \texttt{CC18,TZ}  \\
15    & \texttt{breast-w}                                                          & 699     & 10    & 2  & \texttt{CC18}     \\
16    & \texttt{mfeat-karhunen}                                                    & 2,000   & 65    & 10 & \texttt{CC18}     \\
18    & \texttt{mfeat-morphological}                                               & 2,000   & 7     & 10 & \texttt{CC18}     \\
21    & \texttt{car}                                                               & 1,728   & 7     & 4  & \texttt{TL}       \\
22    & \texttt{mfeat-zernike}                                                     & 2,000   & 48    & 10 & \texttt{CC18,TZ}  \\
23    & \texttt{cmc}                                                               & 1,473   & 10    & 3  & \texttt{CC18}     \\
27    & \texttt{colic}                                                             & 368     & 23    & 2  & \texttt{TZ}       \\
28    & \texttt{optdigits}                                                         & 5,620   & 65    & 10 & \texttt{CC18}     \\
29    & \texttt{credit-approval}                                                   & 690     & 16    & 2  & \texttt{CC18,TZ}  \\
30    & \texttt{page-blocks}                                                       & 5,473   & 11    & 5  & \texttt{TL}       \\
31    & \texttt{credit-g}                                                          & 1,000   & 21    & 2  & \texttt{CC18,TZ}  \\
32    & \texttt{pendigits}                                                         & 10,992  & 17    & 10 & \texttt{CC18}     \\
36    & \texttt{segment}                                                           & 2,310   & 20    & 7  & \texttt{TL}       \\
37    & \texttt{diabetes}                                                          & 768     & 9     & 2  & \texttt{CC18}     \\
38    & \texttt{sick}                                                              & 3,772   & 30    & 2  & \texttt{CC18}     \\
44    & \texttt{spambase}                                                          & 4,601   & 58    & 2  & \texttt{CC18}     \\
46    & \texttt{splice}                                                            & 3,190   & 61    & 3  & \texttt{CC18,TZ}  \\
50    & \texttt{tic-tac-toe}                                                       & 958     & 10    & 2  & \texttt{CC18}     \\
54    & \texttt{vehicle}                                                           & 846     & 19    & 4  & \texttt{CC18,TZ}  \\
60    & \texttt{waveform-5000}                                                     & 5,000   & 41    & 3  & \texttt{TL}       \\
151   & \texttt{electricity}                                                       & 45,312  & 9     & 2  & \texttt{CC18}     \\
179   & \texttt{adult}                                                             & 48,842  & 15    & 2  & \texttt{TL}       \\
180   & \texttt{covertype}                                                         & 110,393 & 55    & 7  & \texttt{TL}       \\
181   & \texttt{yeast}                                                             & 1,484   & 9     & 10 & \texttt{TL}       \\
182   & \texttt{satimage}                                                          & 6,430   & 37    & 6  & \texttt{CC18}     \\
188   & \texttt{eucalyptus}                                                        & 736     & 20    & 5  & \texttt{CC18}     \\
293   & \texttt{covertype}                                                         & 581,012 & 55    & 2  & \texttt{TL}       \\
333   & \texttt{monks-problems-1}                                                  & 556     & 7     & 2  & \texttt{TZ}       \\
458   & \texttt{analcatdata\_authorship}                                           & 841     & 71    & 4  & \texttt{CC18}     \\
469   & \texttt{analcatdata\_dmft}                                                 & 797     & 5     & 6  & \texttt{CC18}     \\
470   & \texttt{profb}                                                             & 672     & 10    & 2  & \texttt{TZ}       \\
554   & \texttt{mnist\_784}                                                        & 70,000  & 785   & 10 & \texttt{CC18}     \\
846   & \texttt{elevators}                                                         & 16,599  & 19    & 2  & \texttt{TZ}       \\
934   & \texttt{socmob}                                                            & 1,156   & 6     & 2  & \texttt{TZ}       \\
999   & \texttt{audiology}                                                         & 226     & 70    & 2  & \texttt{TZ}       \\
1038  & \texttt{gina\_agnostic}                                                    & 3,468   & 971   & 2  & \texttt{TL}       \\
1043  & \texttt{ada\_agnostic}                                                     & 4,562   & 49    & 2  & \texttt{TZ}       \\
1046  & \texttt{mozilla4}                                                          & 15,545  & 6     & 2  & \texttt{TL}       \\
1049  & \texttt{pc4}                                                               & 1,458   & 38    & 2  & \texttt{CC18}     \\
1050  & \texttt{pc3}                                                               & 1,563   & 38    & 2  & \texttt{CC18}     \\
1053  & \texttt{jm1}                                                               & 10,885  & 22    & 2  & \texttt{CC18}     \\
1063  & \texttt{kc2}                                                               & 522     & 22    & 2  & \texttt{CC18}     \\
1067  & \texttt{kc1}                                                               & 2,109   & 22    & 2  & \texttt{CC18,TZ}  \\
1068  & \texttt{pc1}                                                               & 1,109   & 22    & 2  & \texttt{CC18}     \\
1111  & \texttt{KDDCup09\_appetency}                                               & 50,000  & 231   & 2  & \texttt{TL}       \\
1112  & \texttt{KDDCup09\_churn}                                                   & 50,000  & 231   & 2  & \texttt{TL}       \\
1114  & \texttt{KDDCup09\_upselling}                                               & 50,000  & 231   & 2  & \texttt{TL}       \\
1116  & \texttt{musk}                                                              & 6,598   & 168   & 2  & \texttt{TL}       \\
1119  & \texttt{adult-census}                                                      & 32,561  & 16    & 2  & \texttt{TL}       \\
1120  & \texttt{MagicTelescope}                                                    & 19,020  & 12    & 2  & \texttt{TL}       \\
1169  & \texttt{airlines}                                                          & 539,383 & 8     & 2  & \texttt{TZ}       \\
1459  & \texttt{artificial-characters}                                             & 10,218  & 8     & 10 & \texttt{TZ}       \\
1461  & \texttt{bank-marketing}                                                    & 45,211  & 17    & 2  & \texttt{CC18}     \\
1462  & \texttt{banknote-authentication}                                           & 1,372   & 5     & 2  & \texttt{CC18}     \\
1464  & \texttt{blood-transfusion-service-center}                                  & 748     & 5     & 2  & \texttt{CC18}     \\
1467  & \texttt{climate-model-simulation-crashes}                                  & 540     & 21    & 2  & \texttt{TL}       \\
1468  & \texttt{cnae-9}                                                            & 1,080   & 857   & 9  & \texttt{CC18,TZ}  \\
1471  & \texttt{eeg-eye-state}                                                     & 14,980  & 15    & 2  & \texttt{TL}       \\
1475  & \texttt{first-order-theorem-proving}                                       & 6,118   & 52    & 6  & \texttt{CC18}     \\
1476  & \texttt{gas-drift}                                                         & 13,910  & 129   & 6  & \texttt{TL}       \\
1478  & \texttt{har}                                                               & 10,299  & 562   & 6  & \texttt{CC18}     \\
1480  & \texttt{ilpd}                                                              & 583     & 11    & 2  & \texttt{CC18}     \\
1485  & \texttt{madelon}                                                           & 2,600   & 501   & 2  & \texttt{CC18}     \\
1486  & \texttt{nomao}                                                             & 34,465  & 119   & 2  & \texttt{CC18,TZ}  \\
1487  & \texttt{ozone-level-8hr}                                                   & 2,534   & 73    & 2  & \texttt{CC18}     \\
1489  & \texttt{phoneme}                                                           & 5,404   & 6     & 2  & \texttt{CC18}     \\
1494  & \texttt{qsar-biodeg}                                                       & 1,055   & 42    & 2  & \texttt{CC18,TZ}  \\
1497  & \texttt{wall-robot-navigation}                                             & 5,456   & 25    & 4  & \texttt{CC18}     \\
1501  & \texttt{semeion}                                                           & 1,593   & 257   & 10 & \texttt{CC18}     \\
1510  & \texttt{wdbc}                                                              & 569     & 31    & 2  & \texttt{CC18}     \\
1565  & \texttt{heart-h}                                                           & 294     & 14    & 5  & \texttt{TZ}       \\
1590  & \texttt{adult}                                                             & 48,842  & 15    & 2  & \texttt{CC18}     \\
1596  & \texttt{covertype}                                                         & 581,012 & 55    & 7  & \texttt{TL}       \\
4134  & \texttt{Bioresponse}                                                       & 3,751   & 1,777 & 2  & \texttt{CC18,TZ}  \\
4534  & \texttt{PhishingWebsites}                                                  & 11,055  & 31    & 2  & \texttt{CC18}     \\
4538  & \texttt{GesturePhaseSegmentationProcessed}                                 & 9,873   & 33    & 5  & \texttt{CC18,TZ}  \\
6332  & \texttt{cylinder-bands}                                                    & 540     & 40    & 2  & \texttt{CC18}     \\
23381 & \texttt{dresses-sales}                                                     & 500     & 13    & 2  & \texttt{CC18}     \\
23512 & \texttt{higgs}                                                             & 98,050  & 29    & 2  & \texttt{TZ}       \\
23517 & \texttt{numerai28.6}                                                       & 96,320  & 22    & 2  & \texttt{CC18}     \\
40536 & \texttt{SpeedDating}                                                       & 8,378   & 121   & 2  & \texttt{TL}       \\
40646 & \texttt{GAMETES\_Epistasis\_2-Way\_20atts\_0.1H\_EDM-1\_1}                  & 1,600   & 21    & 2  & \texttt{TL}       \\
40647 & \texttt{GAMETES\_Epistasis\_2-Way\_20atts\_0.4H\_EDM-1\_1}                  & 1,600   & 21    & 2  & \texttt{TL}       \\
40648 & \texttt{GAMETES\_Epistasis\_3-Way\_20atts\_0.2H\_EDM-1\_1}                  & 1,600   & 21    & 2  & \texttt{TL}       \\
40649 & \texttt{GAMETES\_Heterogeneity\_20atts\_1600\_Het\_0.4\_0.2\_50\_EDM-2\_001} & 1,600   & 21    & 2  & \texttt{TL}       \\
40650 & \texttt{GAMETES\_Heterogeneity\_20atts\_1600\_Het\_0.4\_0.2\_75\_EDM-2\_001} & 1,600   & 21    & 2  & \texttt{TL}       \\
40668 & \texttt{connect-4}                                                         & 67,557  & 43    & 3  & \texttt{CC18}     \\
40670 & \texttt{dna}                                                               & 3,186   & 181   & 3  & \texttt{CC18}     \\
40680 & \texttt{mofn-3-7-10}                                                       & 1,324   & 11    & 2  & \texttt{TL}       \\
40681 & \texttt{mux6}                                                              & 128     & 7     & 2  & \texttt{TL}       \\
40682 & \texttt{thyroid-new}                                                       & 215     & 6     & 3  & \texttt{TL}       \\
40685 & \texttt{shuttle}                                                           & 58,000  & 10    & 7  & \texttt{TL}       \\
40701 & \texttt{churn}                                                             & 5,000   & 21    & 2  & \texttt{CC18}     \\
40900 & \texttt{Satellite}                                                         & 5,100   & 37    & 2  & \texttt{TL}       \\
40945 & \texttt{Titanic}                                                           & 1,309   & 14    & 2  & \texttt{TL}       \\
40966 & \texttt{MiceProtein}                                                       & 1,080   & 82    & 8  & \texttt{CC18}     \\
40975 & \texttt{car}                                                               & 1,728   & 7     & 4  & \texttt{CC18}     \\
40978 & \texttt{Internet-Advertisements}                                           & 3,279   & 1,558 & 2  & \texttt{CC18}     \\
40979 & \texttt{mfeat-pixel}                                                       & 2,000   & 241   & 10 & \texttt{CC18}     \\
40981 & \texttt{Australian}                                                        & 690     & 15    & 2  & \texttt{TZ}       \\
40982 & \texttt{steel-plates-fault}                                                & 1,941   & 28    & 7  & \texttt{CC18}     \\
40983 & \texttt{wilt}                                                              & 4,839   & 6     & 2  & \texttt{CC18}     \\
40984 & \texttt{segment}                                                           & 2,310   & 20    & 7  & \texttt{CC18}     \\
40994 & \texttt{climate-model-simulation-crashes}                                  & 540     & 21    & 2  & \texttt{CC18}     \\
41027 & \texttt{jungle\_chess\_2pcs\_raw\_endgame\_complete}                        & 44,819  & 7     & 3  & \texttt{CC18,TZ}  \\
41138 & \texttt{APSFailure}                                                        & 76,000  & 171   & 2  & \texttt{TL}       \\
41143 & \texttt{jasmine}                                                           & 2,984   & 145   & 2  & \texttt{TZ}       \\
41147 & \texttt{albert}                                                            & 425,240 & 79    & 2  & \texttt{TZ}       \\
41150 & \texttt{MiniBooNE}                                                         & 130,064 & 51    & 2  & \texttt{TZ}       \\
43945 & \texttt{electricity}                                                       & 38,474  & 9     & 2  & \texttt{TZ}       \\
43973 & \texttt{phoneme}                                                           & 3,172   & 6     & 2  & \texttt{TZ}       \\
46905 & \texttt{Amazon\_employee\_access}                                          & 32,769  & 10    & 2  & \texttt{TA}       \\
46906 & \texttt{anneal}                                                            & 898     & 39    & 5  & \texttt{TA}       \\
46908 & \texttt{APSFailure}                                                        & 76,000  & 171   & 2  & \texttt{TA}       \\
46910 & \texttt{bank-marketing}                                                    & 45,211  & 14    & 2  & \texttt{TA}       \\
46911 & \texttt{Bank\_Customer\_Churn}                                             & 10,000  & 11    & 2  & \texttt{TA}       \\
46912 & \texttt{Bioresponse}                                                       & 3,751   & 1,777 & 2  & \texttt{TA}       \\
46913 & \texttt{blood-transfusion-service-center}                                  & 748     & 5     & 2  & \texttt{TA}       \\
46915 & \texttt{churn}                                                             & 5,000   & 20    & 2  & \texttt{TA}       \\
46916 & \texttt{coil2000\_insurance\_policies}                                     & 9,822   & 86    & 2  & \texttt{TA}       \\
46918 & \texttt{credit-g}                                                          & 1,000   & 21    & 2  & \texttt{TA}       \\
46919 & \texttt{credit\_card\_clients\_default}                                    & 30,000  & 24    & 2  & \texttt{TA}       \\
46920 & \texttt{customer\_satisfaction\_in\_airline}                               & 129,880 & 22    & 2  & \texttt{TA}       \\
46921 & \texttt{diabetes}                                                          & 768     & 9     & 2  & \texttt{TA}       \\
46922 & \texttt{Diabetes130US}                                                     & 71,518  & 48    & 2  & \texttt{TA}       \\
46924 & \texttt{E-CommereShippingData}                                             & 10,999  & 11    & 2  & \texttt{TA}       \\
46927 & \texttt{Fitness\_Club}                                                     & 1,500   & 7     & 2  & \texttt{TA}       \\
46929 & \texttt{GiveMeSomeCredit}                                                  & 150,000 & 11    & 2  & \texttt{TA}       \\
46930 & \texttt{hazelnut-spread-contaminant-detection}                             & 2,400   & 31    & 2  & \texttt{TA}       \\
46932 & \texttt{heloc}                                                             & 10,459  & 24    & 2  & \texttt{TA}       \\
46933 & \texttt{hiva\_agnostic}                                                    & 3,845   & 1,618 & 3  & \texttt{TA}       \\
46935 & \texttt{HR\_Analytics\_Job\_Change\_of\_Data\_Scientists}                   & 19,158  & 13    & 2  & \texttt{TA}       \\
46937 & \texttt{in\_vehicle\_coupon\_recommendation}                               & 12,684  & 25    & 2  & \texttt{TA}       \\
46938 & \texttt{Is-this-a-good-customer}                                           & 1,723   & 14    & 2  & \texttt{TA}       \\
46939 & \texttt{kddcup09\_appetency}                                               & 50,000  & 213   & 2  & \texttt{TA}       \\
46940 & \texttt{Marketing\_Campaign}                                               & 2,240   & 26    & 2  & \texttt{TA}       \\
46941 & \texttt{maternal\_health\_risk}                                            & 1,014   & 7     & 3  & \texttt{TA}       \\
46947 & \texttt{online\_shoppers\_intention}                                       & 12,330  & 18    & 2  & \texttt{TA}       \\
46950 & \texttt{polish\_companies\_bankruptcy}                                     & 5,910   & 65    & 2  & \texttt{TA}       \\
46952 & \texttt{qsar-biodeg}                                                       & 1,054   & 42    & 2  & \texttt{TA}       \\
46955 & \texttt{SDSS17}                                                            & 78,053  & 12    & 3  & \texttt{TA}       \\
46956 & \texttt{seismic-bumps}                                                     & 2,584   & 16    & 2  & \texttt{TA}       \\
46958 & \texttt{splice}                                                            & 3,190   & 61    & 3  & \texttt{TA}       \\
46960 & \texttt{students\_dropout\_and\_academic\_success}                         & 4,424   & 37    & 3  & \texttt{TA}       \\
46962 & \texttt{taiwanese\_bankruptcy\_prediction}                                 & 6,819   & 95    & 2  & \texttt{TA}       \\
46963 & \texttt{website\_phishing}                                                 & 1,353   & 10    & 3  & \texttt{TA}       \\
46969 & \texttt{NATICUSdroid}                                                      & 7,491   & 87    & 2  & \texttt{TA}       \\
46979 & \texttt{jm1}                                                               & 10,885  & 22    & 2  & \texttt{TA}       \\
46980 & \texttt{MIC}                                                               & 1,699   & 112   & 8  & \texttt{TA}       \\
\end{longtable}
\endgroup

\end{document}